\pdfoutput=1
\documentclass[letterpaper, 10pt, conference]{ieeeconf}
\IEEEoverridecommandlockouts   % needed for \thanks
\usepackage[T1]{fontenc}       % Type-1 fonts (no Type-3/bitmapped)
\usepackage{mathptmx}          % Times (IEEE Base-14 Type-1 face), text + math
\usepackage[utf8]{inputenc}
\usepackage{graphicx}
\usepackage{amsmath,amssymb}
\usepackage{bm}                % robust bold math (bold Greek: mathptmx lacks a bold Greek series)
\usepackage{booktabs}
\usepackage{array}
\usepackage{multirow}
\usepackage{siunitx}
\usepackage{xcolor}
\usepackage{cite}
\usepackage[hidelinks,bookmarks=false]{hyperref}
\graphicspath{{figures/}}
\title{\LARGE \bf
Learning Which Correspondences to Trust:\\
Confidence-Weighted Event-Camera Localization in LiDAR Maps}

\author{Panagiotis Kiousis$^{1,2}$, Kuangyi Chen$^{2}$, Jun Zhang$^{2}$, Friedrich Fraundorfer$^{2}$%
  \thanks{$^{1}$Dept.\ of Electrical \& Computer Engineering, Univ. of Patras, Greece.}%
  \thanks{$^{2}$Institute of Visual Computing, Graz University of Technology, Graz, Austria. E-mail: panagiotis.kiousis@student.tugraz.at, \{kuangyi.chen, jun.zhang, friedrich.fraundorfer\}\allowbreak@tugraz.at. Code (CELL): \url{https://github.com/panagiotisq/CELL}.}%
}

\begin{document}
\maketitle
\vspace{-8pt}
\thispagestyle{empty}
\pagestyle{empty}

\begin{abstract}
Localizing an event camera against a pre-built LiDAR map can be cast as dense optical-flow estimation
between a rendered depth view and an event image, followed by a Perspective-$n$-Point (PnP) solver over
the induced 3D--2D correspondences. Existing pipelines rely on geometric consensus during pose estimation, but do not explicitly model the reliability or pose informativeness, i.e., how strongly a correspondence constrains the camera pose, of individual correspondences. We show that the natural way to learn it---using the per-correspondence error to constrain the learning of confidence---suffers from a depth-dependent bias: small pixel errors reside predominantly at large depths and do not lead to high pose informativeness. Instead, in our method (CELL), we learn
a per-correspondence \emph{confidence} end-to-end \emph{through the pose}, using a differentiable
probabilistic PnP whose log-partition term encourages weight configurations that yield a better-constrained pose distribution. The
learned confidence is used in three ways: (i) it reweights the flow supervision in a decoupled training scheme that keeps pose gradients out of the flow/edge backbone; (ii) it drives a probabilistic
correspondence selection at test time; and (iii) together with the network's edge-probability it weights
a final edge-matching refinement. We further design a partial-completion depth representation that
adds signal without hallucinating across large gaps. On M3ED and DSEC our full system improves over the LEAR baseline on the majority of the evaluated sequences: it reduces the median translation error by up to $26.9\%$ and the median rotation error
by up to $15.8\%$.
\end{abstract}

\section{Introduction}
\begin{figure*}[t]
  \centering
  \includegraphics[width=\textwidth]{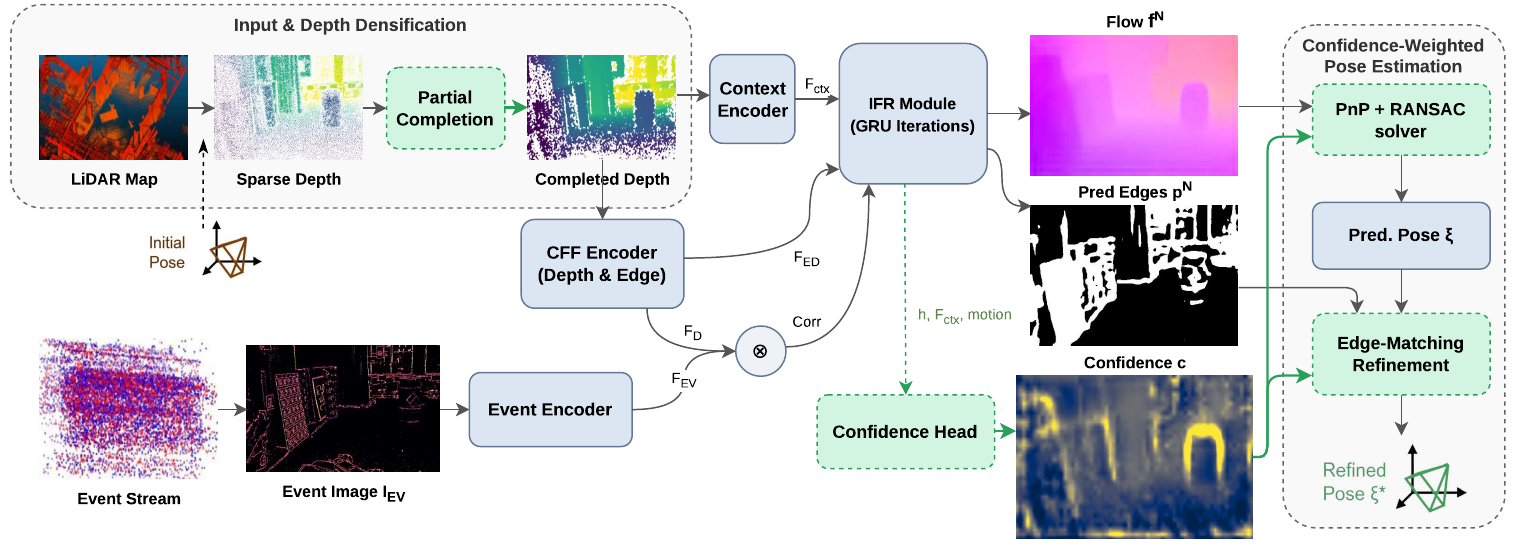}
  \caption{\textbf{Our pipeline.} We extend the LEAR baseline with three components
  (\textcolor[HTML]{28A05A}{\textbf{green, dashed}}): a \emph{partial} depth completion, a
  \emph{confidence head} trained end-to-end through the pose loss, and a \emph{confidence-weighted
  edge-matching refinement}. Sample maps are outputs on \textit{falcon\_indoor\_flight}.}
  \label{fig:pipeline}
\end{figure*}
Event cameras report asynchronous per-pixel brightness changes with microsecond latency and very high
dynamic range, making them attractive for visual perception in fast-motion and challenging-lighting
regimes where frame cameras struggle. A practical form of the problem is \emph{map-relative}
localization: given a pre-built LiDAR point-cloud map, recover the six-degree-of-freedom (6-DoF) pose of an event camera in that
map. Recent methods~\cite{chen2025evloc,chen2026lear} formulate the problem as cross-modal optical-flow
estimation between a rendered depth view and an event image: a sparse depth image is rendered from the
map at a guessed pose, dense flow is estimated between that depth view and an event image with a
RAFT-style recurrent network~\cite{teed2020raft}, the flow is lifted to 3D--2D correspondences, and a pose
is solved with PnP~\cite{persson2018lambdatwist}.

A key limitation is that, beyond rejecting geometrically inconsistent correspondences during pose estimation, the baseline does not explicitly model correspondence reliability or how strongly each correspondence constrains the pose. Structured errors can remain mutually consistent and may survive RANSAC~\cite{fischler1981ransac}-based consensus filtering, while geometrically consistent correspondences can still differ substantially in their usefulness for pose estimation.

A natural strategy is to supervise the correspondence confidence using per-correspondence error~\cite{liu2024egfs} and retain only high-confidence correspondences at inference. However, image-space error is not necessarily well aligned with pose estimation in our setting. Because pixel-space residuals are depth-dependent, distant correspondences can exhibit small flow errors while providing weak constraints on camera translation. As a result, flow-error-supervised confidence can favor correspondences that appear accurate in image space but are not necessarily the most informative for pose estimation, as we show in \mbox{Sec.~\ref{sec:errsup-abl}}. We therefore learn the confidence directly through the pose using differentiable probabilistic PnP~\cite{chen2022epropnp}. Instead of matching confidence to an image-space error target, the probabilistic pose objective optimizes the correspondence weights according to how well they constrain the pose distribution. The resulting confidence is therefore learned with respect to the pose-estimation objective rather than using flow error as a proxy for correspondence quality.

\smallskip\noindent Our contributions are summarized as follows:
\begin{enumerate}
  \item We propose a pose-supervised correspondence confidence learned through differentiable
        probabilistic PnP, together with a decoupled training strategy that uses the learned confidence
        to reweight flow supervision without propagating pose gradients into the backbone.
  \item We use the learned confidence for probabilistic correspondence selection and
        confidence-weighted edge-matching refinement, and introduce partial depth completion to increase useful
	correspondences while preserving geometric boundaries.
  \item Experiments on M3ED and DSEC demonstrate improvements over the LEAR baseline on the
      majority of the evaluated sequences, with gains in translation, rotation, and localization
      accuracy.
\end{enumerate}

\section{Related Work}

\noindent\textbf{Event-based localization.}
Existing event-based localization methods can be broadly grouped into place recognition, direct pose regression, and map-based localization. Place-recognition approaches retrieve the closest entry from a reference database using event representations such as reconstructed images~\cite{fischer2020event}, edge maps~\cite{lee2021eventvlad}, sparse informative pixels~\cite{fischer2022many}, or learned descriptors~\cite{kong2022event}. These methods provide coarse localization but do not directly recover a metric 6-DoF pose. Pose-regression methods instead predict the camera pose directly from event data~\cite{nguyen2019real,jin20216,lin20226,ren2024simple}, but typically encode scene-specific information in the learned model, which can limit generalization to unseen environments. Map-based approaches estimate the pose with respect to an explicit 3D representation, either geometrically by aligning events with semi-dense maps~\cite{yuan2024evit,zuo2024cross}, or photometrically by matching events against appearance predicted from photometric maps~\cite{gallego2017event,bryner2019event} or Gaussian-splatting representations~\cite{liu2024gs}. LiDAR maps, in contrast, provide dense metric geometry unaffected by illumination, motivating event-camera localization directly against LiDAR maps.

\noindent\textbf{Event-to-LiDAR localization.}
Cross-modal event--LiDAR matching is comparatively underexplored. Early efforts target
extrinsic calibration---marker-based~\cite{ta2023l2e} or target-free from edge
correspondences~\cite{xing2023target}---but under controlled conditions that limit scalability, while
E2PNet~\cite{lin2023e2pnet} learns an image-like event representation for point-cloud registration but
reports only coarse accuracy. Closest to this work, EVLoc~\cite{chen2025evloc} casts localization as
dense event--depth flow estimation with a RAFT-style network~\cite{teed2020raft} and lifts the flow to
3D--2D correspondences for a PnP solver; LEAR~\cite{chen2026lear} augments this with a jointly-trained
depth-edge detector and cross-task fusion, making the rendered depth more ``event-like'' and improving robustness. We build directly on LEAR. Its final pose estimation already uses PnP with RANSAC to reject geometrically inconsistent correspondences, but correspondence quality is not explicitly learned before the geometric estimation. In particular, RANSAC inlier rejection does not distinguish between correspondences that are geometrically consistent yet differ substantially in how strongly they constrain the camera pose. Our work therefore complements, rather than replaces, geometric consensus by learning a pose-supervised correspondence confidence.

\noindent\textbf{Confidence-based correspondence learning.}
Down-weighting unreliable correspondences before a geometric solve is well studied outside the event
domain. NG-RANSAC learns sampling weights for RANSAC~\cite{brachmann2019ngransac}; DROID-SLAM predicts
per-pixel confidence weights consumed by a differentiable bundle adjustment~\cite{teed2021droid}; and
EPro-PnP makes PnP itself differentiable through a probabilistic formulation, learning 2D--3D
correspondence weights end-to-end via a pose loss~\cite{chen2022epropnp}. A related strategy uses reprojection error as a supervision or guidance signal; for example, EGFS~\cite{liu2024egfs} uses low-error regions to guide scene-coordinate regression. We observe a limitation of using image-space correspondence error as the supervision signal for confidence learning: pixel-space error is inherently depth-dependent, so low-error correspondences can be biased toward distant, low-parallax points whose small image-space residuals do not necessarily imply strong pose constraints, particularly for translation. We therefore learn confidence directly through a probabilistic pose objective rather than using correspondence error alone as the confidence supervision target.

\section{Preliminaries}
\label{sec:prelim}
We build directly on LEAR~\cite{chen2026lear}, which
casts event-to-LiDAR localization as dense flow estimation followed by a PnP solver. This section summarizes the baseline components and notation required by
our method.

\noindent\textbf{Problem and inputs.} Given a pre-built LiDAR map and a coarse initial pose
$\mathbf{T}_{\text{init}}$, the goal is to recover the camera's 6-DoF pose. The map points are
transformed into the camera frame and projected with a pinhole model at $\mathbf{T}_{\text{init}}$ to
render a sparse depth image $I_D$; in parallel, the events within a short time window are filtered
and accumulated into an event image $I_{EV}$.

\noindent\textbf{Edge-aware flow network.} The depth image is encoded into the depth feature $\mathbf{F}_D$, while a separate event encoder produces the event feature $\mathbf{F}_{EV}$. A context encoder additionally provides context features $\mathbf{F}_{ctx}$. Based on the correlation between $\mathbf{F}_D$ and $\mathbf{F}_{EV}$, an iterative feature refinement module (IFR) progressively updates the flow using a gated recurrent unit (GRU). During refinement, the IFR maintains a hidden state $\mathbf{h}$ and motion features derived from the correlation lookup and the current flow. The recurrent updates produce a sequence of flow estimates $\mathbf{f}^{1:N}$ and edge-probability maps $\mathbf{p}^{1:N}$.

\noindent\textbf{From flow to pose.} For each pixel $\mathbf{u}_i$ with valid depth, the depth value is deprojected to a 3D map point $\mathbf{X}_i$. The final flow prediction provides the corresponding 2D observation in the event image, $\mathbf{x}_i=\mathbf{u}_i+\mathbf{f}^{N}(\mathbf{u}_i)$, where $\mathbf{u}_i=\pi(\mathbf{X}_i,\mathbf{T}_{\text{init}})$ and $\pi(\cdot)$ denotes camera projection. The resulting correspondence set is passed to a PnP solver within a RANSAC scheme to estimate the camera pose.
Although RANSAC rejects geometrically inconsistent correspondences, the
baseline does not explicitly model correspondence-level reliability or
how strongly each correspondence constrains the pose before geometric
estimation. In addition, the sparse projected depth limits the spatial
support from which valid correspondences can be formed.

\section{Method}
\label{sec:method}
Building on the LEAR pipeline described in Sec.~\ref{sec:prelim}, our method centers on learning a pose-supervised confidence for each 3D--2D correspondence. The learned confidence is used to reweight flow supervision during training, to guide correspondence selection at inference, and to weight a subsequent edge-matching refinement---exploiting an edge-probability map the baseline predicts but only ever uses internally for
flow fusion. In addition, we employ a conservative partial depth completion to increase the spatial support of the sparse LiDAR projection while preserving depth discontinuities. An overview of the complete pipeline is shown in Fig.~\ref{fig:pipeline}.

\subsection{Partial depth completion}
\label{sec:partial}
The rendered depth $I_D$ is sparse (approximately 20\% of pixels). Sparse projected depth limits the number and spatial coverage of valid correspondences. However, aggressive completion may propagate depth across large unsupported regions, blurring depth discontinuities. We therefore complete the depth only \emph{partially}---a single dilation of the valid points with a $3\times3$ cross kernel, without filling large gaps.
This increases the fraction of valid depth pixels from approximately 20\% to 68\%---supplying the flow and pose supervision with
far more valid 2D--3D correspondences---while preserving the object edges the edge branch and later edge
matching rely on.

\subsection{Per-correspondence confidence head}
\label{sec:head}
We attach a small head that predicts a dense per-pixel confidence map $\mathbf{c}$, evaluated once per frame. It
reads three feature groups from the IFR: (i) $K{=}4$ snapshots of the GRU hidden state $\mathbf{h}$ (the recurrent
flow estimator's evolving internal state), taken at fixed iteration indices and concatenated; (ii) the context features $\mathbf{F}_{ctx}$; and
(iii) motion features derived from the correlation lookup and current flow. These are concatenated ($4{\cdot}128{+}256{+}128=896$ channels at $1/8$ resolution) into a per-pixel
feature vector $\mathbf{g}$ and passed through a two-layer $1{\times}1$ convolutional
MLP:
\begin{equation}
\mathrm{ConfHead}(\mathbf{g})=\mathrm{Conv}_{1\times1}^{256\to1}\!\big(\mathrm{ReLU}(\mathrm{Conv}_{1\times1}^{896\to256}(\mathbf{g}))\big).
\end{equation}
The single-channel logit map $\mathbf{z}$ is bilinearly upsampled to the
input resolution and converted to a positive confidence map: $\mathbf{c}=\mathrm{softplus}(\mathbf{z})>0$. We use softplus because the confidence is subsequently used as a positive
relative weight rather than a bounded probability. Since the input features
already aggregate spatial and recurrent context, the confidence head itself
remains lightweight and per-pixel. The input features to the confidence head are detached, so the gradients from the confidence branch do not propagate into the flow/edge backbone. The per-correspondence confidence $c_i$ is obtained by sampling $\mathbf{c}$ at the pixel of
correspondence $i$.

\subsection{Learning confidence through the pose}
\label{sec:train}
\begin{figure*}[t]
  \centering
  \includegraphics[width=\textwidth]{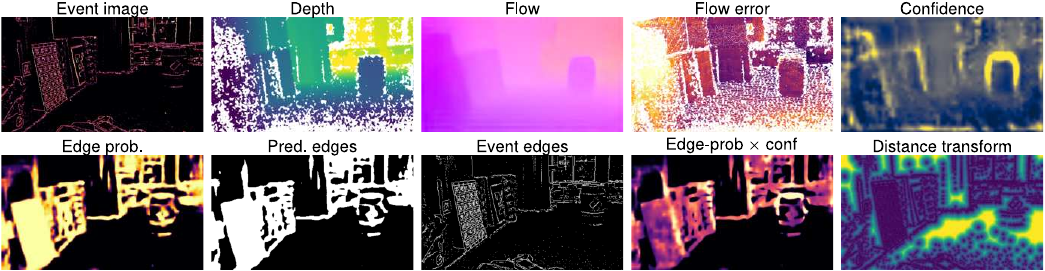}
  \caption{\textbf{Qualitative maps} (\textit{falcon\_indoor\_flight}). \emph{Top:} the flow error
(purple\,$\to$\,yellow $=$ low\,$\to$\,high) is
low almost everywhere, dominated by the depth-dependent perspective effect rather than by pose-relevant
structure---a signal the confidence is \emph{not} trained to reproduce. Instead, supervised only through
the pose loss, the learned confidence reflects both cues: it remains high on some low-error, far-depth
regions, but concentrates most strongly on near-depth, high-parallax object edges---the correspondences
most informative for translation. \emph{Bottom:} the edge-probability, the predicted and event edges, the pointwise product of the edge-probability and confidence maps---the refinement weight (edge-prob $\times$ conf)---and the event-edge distance transform.}
  \label{fig:qual}
  \vspace{-8pt}
\end{figure*}
The natural flow-error-based baseline, following the error-guided selection idea~\cite{liu2024egfs}, learns a
confidence $\kappa_i>0$ by supervising each correspondence with
\begin{equation}
\mathcal{L}^{\text{err}}_i=\kappa_i\tanh\!\big(e_i/\tau\big)-\eta\log\kappa_i,
\label{eq:errsup}
\end{equation}
where $e_i=\lVert\mathbf{f}^{N}(\mathbf{u}_i)-\mathbf{f}_{\text{gt}}(\mathbf{u}_i)\rVert_2$ is the flow error, $\tau$ a
clamping scale, and $\eta$ a regularization weight. This fails here because the supervision signal is strongly
depth-dependent: distant points tend to exhibit smaller pixel-space displacements and residuals. Minimizing this signal therefore induces a preference that is not aligned with pose informativeness, so flow-error-based selection is systematically biased toward distant, low-parallax points (Fig.~\ref{fig:qual}, Table~\ref{tab:errsup}).

\noindent\textbf{Pose-supervised weights.}
We instead learn the weights end-to-end through a differentiable probabilistic PnP~\cite{chen2022epropnp}.
Let $r_i(\mathbf{T})=\rho\big(\lVert\pi(\mathbf{X}_i,\mathbf{T})-\mathbf{x}_i\rVert\big)$ denote the robust
reprojection residual of correspondence $i$, where $\pi$ is the pinhole projection, $\mathbf{x}_i$ is the 2D
observation, and $\rho$ is a Huber kernel. The per-correspondence \emph{pose weights} $\tilde{c}_i =
M\,\mathrm{softmax}_i(\mathbf{z})$ are a softmax of the logit $\mathbf{z}$ over the $M$ correspondences sampled in a frame, scaled so that
$\sum_i \tilde{c}_i = M$ and the weights average to $1$, matching the normalization EPro-PnP's formulation
requires. The pose weights and the softplus confidence $\mathbf{c}$ are computed from the same logit $\mathbf{z}$: the pose weights are used only in the pose loss, while
$\mathbf{c}$ (Sec.~\ref{sec:head}) reweights the flow loss and is used for selection and
refinement at test time. The pose loss combines a target term and a log-partition term, defined as follows:
\begin{equation}
\mathcal{L}_{\text{tgt}} = \textstyle\sum_i \tilde{c}_i\,r_i(\mathbf{T}_{\text{gt}}),
\label{eq:ltgt}
\end{equation}
\begin{equation}
\mathcal{L}_{\text{pred}} = \log\!\int\!\exp\!\big(-\textstyle\sum_i \tilde{c}_i\,r_i(\mathbf{T})\big)\,d\mathbf{T},
\label{eq:lpred}
\end{equation}
\begin{equation}
\mathcal{L}_{\text{pose}} = \mathcal{L}_{\text{tgt}} + \gamma\,\mathcal{L}_{\text{pred}}.
\label{eq:epropnp}
\end{equation}
The target term $\mathcal{L}_{\text{tgt}}$ (weighted residual at
$\mathbf{T}_{\text{gt}}$) alone rewards correspondences with low reprojection error, which tend to be distant, low-parallax points---the source of the depth bias. The
log-partition $\mathcal{L}_{\text{pred}}$ is small only when the weighted correspondences make the pose
distribution \emph{sharp}. We up-weight the log-partition term with $\gamma{=}1.5$ to emphasize pose-level constraint quality, and use the resulting confidence for correspondence selection in Sec.~\ref{sec:select}. The differentiable solver is required only for training, where it makes the
pose loss backpropagate into the confidence head. At inference differentiability is unneeded, and we found
the baseline's standard PnP$+$RANSAC solver to be more accurate and robust. We therefore use it for all test-time results, and the EPro-PnP layer is training-only.

\noindent\textbf{Decoupled training.}
The confidence head is trained together with the flow/edge network but decoupled from it. The pose loss is
applied with the flow detached, so it trains \emph{only} the confidence head (whose inputs are
detached); the flow and edge branches keep training on their own losses. The same
confidence, \emph{detached}, reweights the flow supervision:
\begin{equation}
\mathcal{L}_{\text{flow}}^{c}=\frac{\sum_{u,v}\mathrm{sg}\!\big(\mathbf{c}(u,v)\big)\,\mathbf{1}[\mathbf{f}_{\text{gt}}(u,v)\neq0]\,\lVert\mathbf{f}(u,v)-\mathbf{f}_{\text{gt}}(u,v)\rVert_2}{\sum_{u,v}\mathrm{sg}\!\big(\mathbf{c}(u,v)\big)\,\mathbf{1}[\mathbf{f}_{\text{gt}}(u,v)\neq0]+\epsilon},
\label{eq:reweight}
\end{equation}
a confidence-weighted mean of the per-pixel end-point error ($\ell_2$ norm) over the valid pixels
($\mathrm{sg}$ = stop-gradient; $\mathbf{1}[\mathbf{f}_{\text{gt}}(u,v)\neq0]$ is $1$ where the ground-truth
flow is valid and $0$ otherwise; $\epsilon$ a small constant for numerical stability). This concentrates flow supervision
on confident, pose-important regions and improves the flow, without the pose gradient entering
the backbone. The total objective is
$\alpha\,\mathcal{L}_{\text{flow}}^{c}+\beta\,\mathcal{L}_{\text{edge}}+\lambda\,\mathcal{L}_{\text{pose}}$
($\alpha{=}1$, $\beta{=}100$, $\lambda{=}0.1$).

\subsection{Confidence-guided selection at test time}
\label{sec:select}
Given the learned confidence, we next consider how it should be used when solving for the pose. One strategy is to keep only the correspondences whose confidence is above the per-frame median and
discard the rest. This helps, but it implicitly treats the confidence as \emph{perfectly} reliable: a hard
cutoff throws away every low-ranked correspondence, so any error in the confidence ordering removes
potentially useful points.

Ideally, each correspondence would be \emph{weighted} by its confidence inside the solver, keeping all
points but letting reliable ones dominate. Since the PnP solver does not support per-correspondence weights, we instead bias the correspondence set through confidence-dependent stochastic sampling. Each correspondence is retained with a probability that increases with its confidence,
\begin{equation}
q_i=\mathrm{floor}+(1-\mathrm{floor})\,\hat{c}_i ,
\label{eq:pweight}
\end{equation}
where $\hat{c}_i\in[0,1]$ is the confidence normalized within the frame (min--max). The raw confidence is a
softplus output of arbitrary scale; the per-frame normalization makes the
probability of sampling well-defined and invariant to that scale. With $\mathrm{floor}{=}0$ (denoted
floor0), the lowest-confidence point is kept with probability $0$ and the highest with probability
$1$, linearly in between---a soft analogue of weighting that biases, rather than truncates, the sampled
subset.

Because the confidence is not perfectly reliable, a purely proportional rule ($\mathrm{floor}{=}0$) can still
discard a useful but low-scored correspondence. We hedge against this with a non-zero floor: with
$\mathrm{floor}{=}0.5$ (denoted floor0.5), even a correspondence with the minimum normalized confidence is retained with
probability $0.5$, rising linearly to $1$ for the most confident. This makes the selection robust to
imperfect confidence while still favouring the reliable correspondences.

The sampling also reduces the number of correspondences passed to the
geometric solver. With the observed average normalized confidence, floor0
and floor0.5 retain approximately 28\% and 64\% of the correspondences,
respectively, compared with 100\% when using all points. This reduces the workload of the subsequent PnP+RANSAC stage.

\begin{table*}[t]
    \centering
    \caption{\textit{Comparison with EVLoc and LEAR on M3ED and DSEC.} Metrics as in Sec.~\ref{sec:setup}; \textbf{Ours}: confidence-guided
    selection with confidence-weighted edge-matching refinement; bold: best per row. EVLoc results as reported in~\cite{chen2026lear} (no accuracy).}
    \label{tab:main}
    \setlength{\tabcolsep}{6pt}
    {
    \begin{tabular}{lll ccc ccc ccc}
        \toprule\toprule
        \multirow{2}{*}{Dataset} & \multirow{2}{*}{Test Scene} & \multirow{2}{*}{Time} & \multicolumn{3}{c}{EVLoc~\cite{chen2025evloc}} & \multicolumn{3}{c}{LEAR} & \multicolumn{3}{c}{Ours}\\
        \cmidrule(lr){4-6}\cmidrule(lr){7-9}\cmidrule(lr){10-12}
        & & & T[cm]$\downarrow$ & R[$^\circ$]$\downarrow$ & Acc$\uparrow$ & T[cm]$\downarrow$ & R[$^\circ$]$\downarrow$ & Acc$\uparrow$ & T[cm]$\downarrow$ & R[$^\circ$]$\downarrow$ & Acc$\uparrow$\\
        \midrule
        \multirow{10}{*}{M3ED}
        & \textit{car\_forest\_into\_ponds} & day & 19.76 & 0.86 & -- & 15.15 & 0.739 & 72.8 & \textbf{12.25} & \textbf{0.711} & \textbf{82.9}\\
        & \textit{car\_urban\_day\_penno} & day & 14.15 & 0.60 & -- & 12.14 & \textbf{0.525} & 89.1 & \textbf{11.94} & 0.530 & \textbf{90.3}\\
        & \textit{falcon\_outdoor\_day\_penno\_parking} & day & 25.11 & 1.32 & -- & 23.61 & 1.186 & 48.8 & \textbf{19.84} & \textbf{1.000} & \textbf{59.2}\\
        & \textit{falcon\_indoor\_flight} & - & 8.11 & 0.97 & -- & 6.95 & 0.796 & 29.3 & \textbf{6.19} & \textbf{0.670} & \textbf{39.1}\\
        & \textit{spot\_indoor\_building} & - & 12.88 & 2.07 & -- & 10.85 & 1.718 & 13.6 & \textbf{10.53} & \textbf{1.672} & \textbf{16.7}\\
        & \textit{spot\_forest\_road} & day & 13.55 & 0.60 & -- & 12.41 & 0.526 & 77.1 & \textbf{9.12} & \textbf{0.522} & \textbf{82.9}\\
        & \textit{falcon\_forest\_into\_forest} & day & 20.46 & 1.87 & -- & \textbf{18.59} & 2.053 & 40.3 & 18.64 & \textbf{1.768} & \textbf{46.5}\\
        & \textit{spot\_outdoor\_day\_penno} & day & 17.89 & 0.88 & -- & 14.81 & 0.713 & 73.5 & \textbf{13.13} & \textbf{0.659} & \textbf{84.7}\\
        & \textit{spot\_outdoor\_night\_penno} & night & 34.85 & 1.71 & -- & 26.94 & 1.210 & 38.2 & \textbf{24.36} & \textbf{1.207} & \textbf{46.7}\\
        \cmidrule{2-12}
        & Average & & 18.53 & 1.21 & -- & 15.72 & 1.052 & 53.6 & \textbf{14.00} & \textbf{0.971} & \textbf{61.0}\\
        \midrule
        \multirow{14}{*}{DSEC}
        & \textit{thun\_00} & dawn/dusk & 8.26 & 0.42 & -- & 8.50 & 0.365 & 100.0 & \textbf{6.21} & \textbf{0.334} & 100.0\\
        & \textit{zurich\_city\_00} & dawn/dusk & 7.94 & 0.37 & -- & 7.15 & \textbf{0.348} & 98.9 & \textbf{6.68} & 0.353 & \textbf{99.6}\\
        & \textit{zurich\_city\_01} & dawn/dusk & 7.63 & 0.30 & -- & 6.68 & \textbf{0.283} & 98.0 & \textbf{6.49} & 0.284 & 98.0\\
        & \textit{zurich\_city\_02} & dawn/dusk & 7.93 & 0.31 & -- & \textbf{7.07} & 0.282 & \textbf{99.6} & 7.35 & \textbf{0.276} & 99.2\\
        & \textit{zurich\_city\_03} & night & 8.79 & 0.61 & -- & 8.60 & 0.554 & 88.7 & \textbf{7.42} & \textbf{0.543} & \textbf{90.9}\\
        & \textit{zurich\_city\_04} & dawn/dusk & 7.49 & 0.30 & -- & 6.56 & 0.282 & \textbf{99.3} & \textbf{5.89} & \textbf{0.259} & 98.6\\
        & \textit{zurich\_city\_05} & day & 6.42 & 0.28 & -- & 5.67 & 0.256 & 99.9 & \textbf{5.41} & \textbf{0.237} & \textbf{100.0}\\
        & \textit{zurich\_city\_06} & day & 8.12 & 0.39 & -- & \textbf{6.69} & \textbf{0.330} & \textbf{97.2} & 6.74 & 0.349 & 97.0\\
        & \textit{zurich\_city\_07} & day & 7.27 & 0.34 & -- & 6.52 & 0.292 & 99.9 & \textbf{6.47} & \textbf{0.291} & 99.9\\
        & \textit{zurich\_city\_08} & dawn/dusk & 6.62 & 0.27 & -- & 5.73 & 0.240 & \textbf{100.0} & \textbf{5.72} & \textbf{0.235} & 99.5\\
        & \textit{zurich\_city\_09} & night & 6.38 & 0.29 & -- & 5.94 & 0.312 & 99.8 & \textbf{5.80} & \textbf{0.271} & \textbf{100.0}\\
        & \textit{zurich\_city\_10} & night & 8.37 & 0.35 & -- & 7.47 & 0.342 & 98.5 & \textbf{7.13} & \textbf{0.325} & \textbf{99.5}\\
        & \textit{zurich\_city\_11} & dawn/dusk & 6.14 & 0.26 & -- & 5.77 & 0.251 & 100.0 & \textbf{5.45} & \textbf{0.214} & 100.0\\
        \cmidrule{2-12}
        & Average & & 7.49 & 0.35 & -- & 6.80 & 0.318 & 98.4 & \textbf{6.37} & \textbf{0.305} & \textbf{98.6}\\
        \bottomrule\bottomrule
    \end{tabular}}
\end{table*}

\subsection{Confidence-weighted edge-matching refinement}
\label{sec:refine}
Finally we refine the pose by aligning the predicted depth-edges to the observed event edges
(Fig.~\ref{fig:qual}, bottom row). Predicted
edges (from the edge-probability map $\mathbf{p}$, thresholded and skeletonized) are deprojected through $I_D$ to
3D points $\mathbf{X}_i$; we minimize a robust chamfer distance to the distance transform $\mathrm{DT}$ of the
event edges over the pose, parameterized as $\bm{\xi}=(\bm{\omega},\bm{t})\in\mathbb{R}^6$
(axis-angle rotation $\bm{\omega}$ and translation $\bm{t}$), with a regularization term to the
initial pose:
\begin{equation}
E(\bm{\xi})=\frac{\sum_i w_i\,\rho\big(\mathrm{DT}[\pi(\mathbf{X}_i,\bm{\xi})]\big)}{\sum_i w_i}
\;+\;a_w\,\|\bm{\xi}-\bm{\xi}_{\text{init}}\|^2,
\label{eq:refine}
\end{equation}
solved with LBFGS, where $\rho$ is the Huber loss, $\pi(\mathbf{X}_i,\bm{\xi})$ projects $\mathbf{X}_i$ under the pose
given by $\bm{\xi}$, $\bm{\xi}_{\text{init}}$ is the initial pose, and $a_w$ weights the regularization term.
Each edge is weighted by
\emph{both} the edge-probability and the confidence, $w_i=p_i\,c_i$: the edge-probability indicates whether a genuine edge is present at that location, while the confidence reflects the correspondence's pose reliability. The combination gives the best accuracy.

\section{Experiments}

\subsection{Setup}
\label{sec:setup}

\noindent\textbf{Datasets.} We evaluate on M3ED~\cite{chaney2023m3ed} and DSEC~\cite{gehrig2021dsec}.
M3ED provides LiDAR maps with ground-truth trajectories and is our primary benchmark,
spanning heterogeneous environments---indoor flight, outdoor driving, and forest. DSEC adds more complex
driving scenarios but offers only disparity images; following LEAR~\cite{chen2026lear}, we reconstruct a
per-sample point cloud by back-projecting the disparity-derived depth through the pinhole camera model.

\noindent\textbf{Data processing.} As in LEAR~\cite{chen2026lear}, a coarse
initial pose is simulated by perturbing the ground truth (the identity for DSEC) with uniform random
offsets of up to $\pm50$\,cm in translation and $\pm5^\circ$ in rotation; the task is to recover the true
pose. Inputs are cropped to fixed per-dataset resolutions---$512\times288$ for M3ED and $480\times360$ for DSEC---and the depth normalization scale (the
local-map depth ceiling) is set per environment ($10$\,m indoor, $100$\,m outdoor, $50$\,m DSEC). At test time, the 6-DoF pose is estimated from the predicted 3D--2D correspondences using PoseLib~\cite{PoseLib,sattler2019ransaclib} with PnP+RANSAC and a 12-pixel reprojection-error inlier threshold. The differentiable EPro-PnP~\cite{chen2022epropnp} layer is used only during training.

\noindent\textbf{Baseline.} Our direct baseline is LEAR~\cite{chen2026lear}, on which our method is built. All shared components, data splits, training schedules, and geometric settings are kept identical to the baseline; differences are restricted to the components introduced in Sec.~\ref{sec:method}.

\noindent\textbf{Training.} For M3ED a separate model is trained per scene, as the dataset spans
heterogeneous environments: for scenes with multiple sequences, the last sequence is held out for
evaluation and the rest are used for training; single-sequence scenes instead use a
temporal $70/30$ train/test split with no overlap. For DSEC, whose scenes share driving dynamics and
sensor configuration, a single model is trained across all scenes: for scenes with multiple sequences, the
final sequence is held out for evaluation and the rest are pooled for training, while single-sequence
scenes are excluded from training entirely and used only for evaluation. Every model is trained for
$100$ epochs with batch size $2$, initial learning rate $4\times10^{-5}$, weight decay $5\times10^{-5}$, a
\emph{OneCycleLR} schedule, and the AdamW optimizer; the IFR module runs $12$ iterations at training and
$24$ at test. Training uses a single NVIDIA RTX~5070~Ti GPU.

\noindent\textbf{Metrics.} We report the \emph{median} translation (cm) and rotation ($^\circ$) errors as
the primary pose metrics, an accuracy rate (the fraction of frames within a pose-error threshold, applied
per environment: $(5\,\text{cm},5^\circ)$ indoors and $(25\,\text{cm},2^\circ)$ outdoors), and the flow
End-Point Error (EPE, px), averaged per frame over the correspondences kept by the selection and reported as its mean (EPE$_\text{mn}$) and median (EPE$_\text{md}$) over frames.

\begin{figure*}[t]
  \centering
  \includegraphics[width=\textwidth]{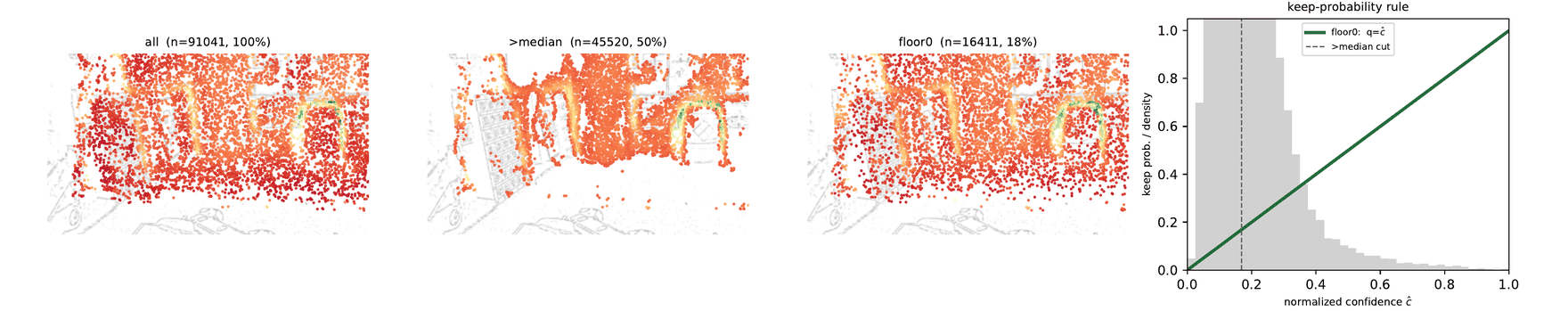}
  \caption{\textbf{Confidence-guided selection} (\textit{falcon\_indoor\_flight}). Correspondences kept by
  each rule, coloured by confidence (green\,=\,high, red\,=\,low; subsampled for clarity), and the
  probability of sampling rule overlaid on the confidence histogram. floor0 keeps a broad but confidence-biased
  subset, whereas the hard \textgreater\,median cut discards half the points.}
  \label{fig:selection}
  \vspace{-8pt}
\end{figure*}
\subsection{Main results}
Table~\ref{tab:main} compares our full system with EVLoc and the LEAR baseline across all 22 evaluated sequences. Our method outperforms EVLoc in both median translation and rotation error on all 22 sequences, by 18.5\% and 14.9\% on average. Compared with LEAR, it reduces the median translation error on 19 sequences and the median rotation error on 18, while improving localization accuracy on 14. Unless otherwise stated, the improvements reported below are relative to LEAR and computed per sequence and then averaged: translation and rotation are reported as relative error reductions, whereas accuracy is reported as an absolute percentage-point gain.

Across the nine M3ED sequences, our method reduces the median translation error by 10.9\% on average, with a maximum reduction of 26.5\%, and reduces the median rotation error by 6.6\% on average, with a maximum reduction of 15.8\%. Localization accuracy improves on all nine sequences, by 7.4 percentage points on average and by up to 11.2 points.

Across the 13 DSEC sequences, our method reduces the median translation error by 5.7\% on average, with a maximum reduction of 26.9\%, and the median rotation error by 4.3\% on average, with a maximum reduction of 14.7\%. Localization accuracy is already high for the LEAR baseline, reaching at least 98\% on 11 of the 13 sequences, leaving limited room for further improvement. Our method nevertheless maintains or improves accuracy on most sequences.

\subsection{Flow-error compared to pose supervision}
\label{sec:errsup-abl}
\begin{table}[t]
    \centering
    \caption{\textit{Flow-error-supervised compared to pose-supervised confidence} (\textit{falcon\_indoor\_flight}, sparse depth): T[\textnormal{cm}]/R[$^\circ$]/Acc/mean depth of kept points [\textnormal{m}].}
    \label{tab:errsup}
    \setlength{\tabcolsep}{27pt}
    {%
    \begin{tabular}{cc}
        \toprule\toprule
        Flow-error-supervised & Pose-supervised \\
        \midrule
        15.69/1.59/4.3/7.04 & 7.74/0.83/22.4/5.92 \\
        \bottomrule\bottomrule
    \end{tabular}}
\end{table}

We construct a flow-error-supervised confidence baseline inspired by
EGFS~\cite{liu2024egfs} (Eq.~\eqref{eq:errsup}), using the same network architecture, training data, and optimization schedule as our pose-supervised model. To expose the difference between the two confidence signals, we retain only the top 10\% most confident correspondences at inference. This deliberately aggressive selection degrades both methods, but the degradation is substantially larger for flow-error-supervised confidence. As shown in Table~\ref{tab:errsup}, flow-error supervision tends to assign higher confidence to more distant correspondences, resulting in markedly worse pose estimation. In contrast, the pose-supervised confidence retains correspondences with lower average depth and preserves substantially better localization.

\subsection{Confidence-guided selection}
\label{sec:select-abl}
\begin{table}[t]
    \centering
    \caption{\textit{Correspondence selection} (non-refined poses, $22$ sequences). W/N: sequences on which a rule
    beats the baseline; $\Delta$: mean change (median T/R in \%, Acc in points).}
    \label{tab:selection}
    \resizebox{\columnwidth}{!}{%
    \begin{tabular}{l cc cc cc}
        \toprule\toprule
        \multirow{2}{*}{Selection}
          & \multicolumn{2}{c}{med\_t} & \multicolumn{2}{c}{med\_r} & \multicolumn{2}{c}{Acc} \\
        & W/N$\uparrow$ & $\Delta$\%$\downarrow$ & W/N$\uparrow$ & $\Delta$\%$\downarrow$ & W/N$\uparrow$ & $\Delta$pp$\uparrow$ \\
        \midrule
        all                  & 18/22 & $-$7.1 & 13/22 & $-$2.1 & \textbf{14/22} & +2.4 \\
        floor0               & 18/22 & \textbf{$-$7.5} & \textbf{15/22} & \textbf{$-$3.2} & \textbf{14/22} & \textbf{+2.8} \\
        floor0.5             & \textbf{19/22} & $-$7.2 & 13/22 & $-$2.5 & \textbf{14/22} & +2.5 \\
        \textgreater median  & 14/22 & $-$5.2 & 14/22 & $-$3.0 & 13/22 & +2.4 \\
        \bottomrule\bottomrule
    \end{tabular}}
\end{table}

Table~\ref{tab:selection} compares the four selection rules against the baseline on the \emph{non-refined}
poses (before edge-matching refinement), isolating the effect of selection. The \emph{all} row is our model without any
selection: it already accounts for most of the improvement over the baseline, a gain that stems from the partial
depth and the decoupled confidence training. Selection operates on top of this model at essentially no runtime
cost (Table~\ref{tab:runtime}) and adds a further, smaller gain. Pooled over all $22$ sequences, floor0 gives the best overall results, being best or tied on five
of the six columns, with the largest mean reductions in both translation ($7.5\%$) and rotation ($3.2\%$) error. The hard
\textgreater median rule remains competitive on rotation but falls behind on translation: discarding half of the correspondences removes points the solver needs. On DSEC, where the
baseline is already near-saturated, selection is largely neutral (all\,$\approx$\,floor0\,$\approx$\,floor0.5)
and only the hard cut hurts. Fig.~\ref{fig:selection} visualises what each rule keeps.

\begin{figure*}[t]
  \centering
  \includegraphics[width=\textwidth]{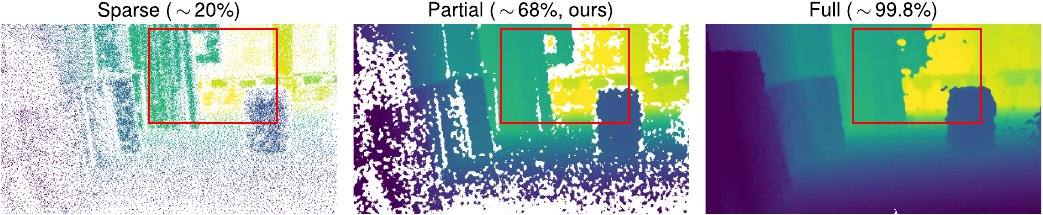}
  \caption{\textbf{Depth-completion variants} (on \textit{falcon\_indoor\_flight}): sparse LiDAR
  ($\sim$20\%), our edge-preserving \emph{partial} completion ($\sim$68\%), and \emph{full} completion
  ($\sim$99.8\%). \emph{Full} propagates depth across object boundaries (red box), unlike partial
  (Table~\ref{tab:depth}).}
  \label{fig:depth-full}
\end{figure*}
\subsection{Edge-matching refinement weighting}
\label{sec:refine-abl}
The refinement of Eq.~\eqref{eq:refine} weights each predicted edge by $w_i$. We compare three choices:
\textbf{none} (uniform, $w_i{=}1$); \textbf{conf} (the per-correspondence confidence $c_i$ sampled at the
edge pixel); and \textbf{both} ($w_i{=}p_i\,c_i$, edge-probability $\times$ confidence). We report the mean accuracy over all sequences.

\begin{table}[t]
    \centering
    \caption{\textit{Edge weighting for the refinement}: mean accuracy and its gain over LEAR and the non-refined
    pose.}
    \label{tab:refine-weight}
    \resizebox{\columnwidth}{!}{%
    \begin{tabular}{lcccc}
        \toprule\toprule
        Edge weighting & Mean Acc & $\Delta$ vs LEAR & $\Delta$ vs non-ref. & $\geq$ base \\
        \midrule
        none (uniform)                 & 83.18 & +3.06 & +0.26 & 16/22 \\
        conf (confidence)         & 83.21 & +3.09 & +0.29 & 16/22 \\
        \textbf{both} (edge-prob $\times$ conf) & \textbf{83.24} & \textbf{+3.12} & \textbf{+0.32} & 16/22 \\
        \bottomrule\bottomrule
    \end{tabular}}
\end{table}

As Table~\ref{tab:refine-weight} shows, all three weightings improve accuracy over both the baseline and the
non-refined pose. Weighting by the confidence provides a small additional improvement over uniform weighting (+0.29 vs. +0.26 points relative to the non-refined pose), suggesting that the learned confidence can also provide a useful weighting cue for the refinement. Weighting by
\textbf{both} the edge-probability and the confidence is strongest overall ($+3.12$/$+0.32$). The edge probability and correspondence confidence provide complementary weighting cues: the former reflects the predicted presence of an edge, whereas the latter is learned through pose supervision. Multiplying the two keeps edges that are \emph{both} geometrically reliable and genuinely present,
which is what the chamfer term needs. We therefore adopt the \textbf{both} weighting.

\subsection{Depth completion: sparse vs.\ partial vs.\ full}
\label{sec:partial-abl}
Table~\ref{tab:depth} isolates the depth input by comparing the three completions on our model trained with the decoupled scheme (Sec.~\ref{sec:train}), at the \emph{all} selection (no confidence selection), on a data-rich outdoor scene. A \emph{full} completion is
counter-productive: it fills nearly every pixel ($\sim$99.8\%) but hallucinates depth at object margins
(Fig.~\ref{fig:depth-full}), which \emph{adds geometrically wrong correspondences}---accuracy drops below even the sparse input
($50.6$ vs.\ $51.1$) and the flow itself degrades (median EPE $6.06\!\to\!6.42$\,px, with the mean rising
more sharply, $8.74\!\to\!10.23$). The \emph{partial}
completion instead fills conservatively ($\sim$68\%): it grows the valid regions enough to add useful
correspondences while preserving object boundaries, and is the best of the three on every metric
($+6.5$ accuracy, $-2.4$\,cm median translation, and lowest EPE).

\begin{table}[t]
    \centering
    \caption{\textit{Depth-input completion} on \textit{falcon\_outdoor\_day\_penno\_parking} (all
    correspondences, no selection).}
    \label{tab:depth}
    \resizebox{\columnwidth}{!}{%
    \begin{tabular}{lccccc}
        \toprule\toprule
        Depth input & Valid & med\_t\,[cm]$\downarrow$ & acc$_\text{O}$\,[\%]$\uparrow$
          & EPE$_\text{mn}$$\downarrow$ & EPE$_\text{md}$$\downarrow$ \\
        \midrule
        sparse                    & $\sim$20\%   & 23.15 & 51.1 & 8.74 & 6.06 \\
        full dense                & $\sim$99.8\% & 23.26 & 50.6 & 10.23 & 6.42 \\
        \textbf{partial (ours)}   & $\sim$68\%   & \textbf{20.80} & \textbf{57.6} & \textbf{8.08} & \textbf{5.78} \\
        \bottomrule\bottomrule
    \end{tabular}}
\end{table}

\subsection{Incremental ablation}
\label{sec:ablation}
\begin{table}[t]
    \centering
    \caption{\textit{Incremental ablation} on \textit{falcon\_outdoor\_day\_penno\_parking}: each row adds one
    component; parentheses give the change vs.\ the previous row. $^\dagger$Refinement changes only the pose, so
    EPE equals the floor0 row.}
    \label{tab:ablation}
    \resizebox{\columnwidth}{!}{%
    \begin{tabular}{lccccc}
        \toprule\toprule
        Configuration & med\_t\,$\downarrow$ & med\_r\,$\downarrow$ & Acc\,$\uparrow$ & EPE$_\text{mn}$\,$\downarrow$ & EPE$_\text{md}$\,$\downarrow$ \\
        \midrule
        base: LEAR (sparse)               & 23.61 & 1.186 & 48.8 & 9.42 & 5.97 \\
        $+$ partial depth           & 22.65 {\scriptsize($-$0.96)} & 1.240 {\scriptsize(+0.054)} & 51.1 {\scriptsize(+2.3)} & 9.63 & 6.31 \\
        $+$ decoupled conf.\ training     & 20.80 {\scriptsize($-$1.85)} & 1.080 {\scriptsize($-$0.160)} & 57.6 {\scriptsize(+6.5)} & 8.08 & 5.78 \\
        $+$ confidence selection          & 20.06 {\scriptsize($-$0.74)} & 1.027 {\scriptsize($-$0.053)} & 58.9 {\scriptsize(+1.3)} & 7.48 & 5.27 \\
        $+$ edge-matching refinement (\textbf{Ours}) & 19.84 {\scriptsize($-$0.22)} & 1.000 {\scriptsize($-$0.027)} & 59.2 {\scriptsize(+0.3)} & \multicolumn{2}{c}{= floor0$^\dagger$} \\
        \midrule
        net (base\,$\to$\,Ours)           & \textbf{19.84} {\scriptsize($-$3.77)} & \textbf{1.000} {\scriptsize($-$0.186)} & \textbf{59.2} {\scriptsize(+10.4)} & 7.48 & 5.27 \\
        \bottomrule\bottomrule
    \end{tabular}}
\end{table}

Table~\ref{tab:ablation} incrementally adds the proposed components on \textit{falcon\_outdoor\_day\_penno\_parking}.
Partial depth completion and decoupled confidence training provide the largest accuracy gains, improving accuracy by 2.3 and 6.5 percentage points, respectively. Confidence-guided selection adds a further 1.3-point improvement, while the final edge-matching refinement provides a smaller additional gain of 0.3 points and reduces the median translation error by 0.22~cm. Overall, the full system reduces the median translation and rotation errors by 3.77~cm and $0.186^\circ$ and improves localization accuracy by 10.4 percentage points over the baseline.

\subsection{Halving the iteration budget}
\label{sec:it12}
\begin{table}[t]
    \centering
    \caption{\textit{Halved iteration budget}: ours at 12 IFR iterations vs.\ the 24-iteration LEAR baseline
    (W/N and $\Delta$ as in Table~\ref{tab:selection}).}
    \label{tab:it12}
    \resizebox{0.85\columnwidth}{!}{%
    \begin{tabular}{l cc cc cc}
        \toprule\toprule
        \multirow{2}{*}{Split}
          & \multicolumn{2}{c}{med\_t} & \multicolumn{2}{c}{med\_r} & \multicolumn{2}{c}{Acc} \\
        & W/N$\uparrow$ & $\Delta$\%$\downarrow$ & W/N$\uparrow$ & $\Delta$\%$\downarrow$ & W/N$\uparrow$ & $\Delta$pp$\uparrow$ \\
        \midrule
        M3ED (9)  & 6/9   & $-$3.2 & 6/9   & $-$1.8 & 7/9   & +3.3 \\
        DSEC (13) & 7/13  & $-$3.2 & 7/13  & $-$1.6 & 5/13  & $-$0.2 \\
        \midrule
        All (22)  & 13/22 & $-$3.2 & 13/22 & $-$1.7 & 12/22 & +1.2 \\
        \bottomrule\bottomrule
    \end{tabular}}
\end{table}

As IFR flow updates dominate the network cost (Table~\ref{tab:runtime}), Table~\ref{tab:it12} runs our full recipe at 12 iterations against the 24-iteration LEAR baseline. At roughly half the flow-update cost, it still reduces the median translation error on 13/22 sequences and increases accuracy on 12/22, mainly on M3ED (7/9 scenes, +3.3 points on average). On near-saturated DSEC, both median errors still improve.

\subsection{Runtime and memory}
\label{sec:runtime}
\begin{table}[t]
    \centering
    \caption{\textit{Runtime and memory} on \textit{falcon\_indoor\_flight}; memory is the peak
    allocation, excluding the fixed CUDA context ($\approx$\,300\,MB).}
    \label{tab:runtime}
    \resizebox{\columnwidth}{!}{%
    \begin{tabular}{l l c c}
        \toprule\toprule
        Method & Stage & Runtime [ms] & Memory [MB] \\
        \midrule
        \multirow{2}{*}{Baseline (LEAR)}
          & Feature encoding            & 8.89   & \multirow{2}{*}{335} \\
          & Flow estimation (24 iters)  & 77.15  & \\
        \midrule
        \multirow{4}{*}{Ours}
          & Feature encoding            & 8.90   & \multirow{4}{*}{339} \\
          & Flow estimation (24/12 iters) & 76.24/39.15 & \\
          & Confidence head             & 0.09   & \\
          & Edge-matching refinement    & 203.82 & \\
        \bottomrule\bottomrule
    \end{tabular}}
\end{table}

Table~\ref{tab:runtime} breaks down the per-frame cost. The core network (feature encoding $+$ IFR flow
estimation) is essentially unchanged from the baseline: the confidence head and the partial-depth
dilation add negligible runtime and memory over the baseline network, so our accuracy gains come at essentially
no extra network cost. Flow estimation dominates the network and is linear in the iteration count, so halving it
($24\!\to\!12$) roughly halves that stage. The edge-matching refinement is the one substantial addition. It is therefore
best viewed as an \emph{optional} final stage: with it disabled, the network's runtime and memory are on par
with the baseline, yet localization still improves over it (Table~\ref{tab:selection}).

\section{Conclusion}
We presented a pose-supervised correspondence confidence for localizing event cameras in LiDAR maps. Learned \emph{through} the pose via a differentiable-PnP loss, the confidence is reused throughout the pipeline: it reweights flow supervision in a decoupled training scheme, drives a probabilistic correspondence selection at test time, and---combined with edge-probability---weights a final edge-matching refinement. Together, these components improve localization over the LEAR baseline on both M3ED and DSEC.

\textbf{Limitations.} The learned confidence needs sufficient training data to become reliable, which limits its benefit on data-scarce scenes, and the gains on DSEC are modest since accuracy there is already near-saturated. The method also depends on the initial pose used to render the depth view and may degrade when this pose is far from the true one.



\begin{thebibliography}{10}
\providecommand{\url}[1]{#1}
\csname url@samestyle\endcsname
\providecommand{\newblock}{\relax}
\providecommand{\bibinfo}[2]{#2}
\providecommand{\BIBentrySTDinterwordspacing}{\spaceskip=0pt\relax}
\providecommand{\BIBentryALTinterwordstretchfactor}{4}
\providecommand{\BIBentryALTinterwordspacing}{\spaceskip=\fontdimen2\font plus
\BIBentryALTinterwordstretchfactor\fontdimen3\font minus
  \fontdimen4\font\relax}
\providecommand{\BIBforeignlanguage}[2]{{%
\expandafter\ifx\csname l@#1\endcsname\relax
\typeout{** WARNING: IEEEtran.bst: No hyphenation pattern has been}%
\typeout{** loaded for the language `#1'. Using the pattern for}%
\typeout{** the default language instead.}%
\else
\language=\csname l@#1\endcsname
\fi
#2}}
\providecommand{\BIBdecl}{\relax}
\BIBdecl

\bibitem{chen2025evloc}
K.~Chen, J.~Zhang, and F.~Fraundorfer, ``{EVLoc}: Event-based visual
  localization in {LiDAR} maps via event-depth registration,'' in \emph{Proc.
  IEEE Int. Conf. Robot. Autom. (ICRA)}, 2025.

\bibitem{chen2026lear}
K.~Chen, J.~Zhang, Y.~Hu, Y.~Zhou, and F.~Fraundorfer, ``{LEAR}: Learning
  edge-aware representations for event-to-{LiDAR} localization,'' in
  \emph{Proc. IEEE Int. Conf. Robot. Autom. (ICRA)}, 2026.

\bibitem{teed2020raft}
Z.~Teed and J.~Deng, ``{RAFT}: Recurrent all-pairs field transforms for optical
  flow,'' in \emph{Proc. Eur. Conf. Comput. Vis. (ECCV)}, 2020, pp. 402--419.

\bibitem{persson2018lambdatwist}
M.~Persson and K.~Nordberg, ``Lambda twist: An accurate fast robust perspective
  three point ({P3P}) solver,'' in \emph{Proc. Eur. Conf. Comput. Vis.}, 2018.

\bibitem{fischler1981ransac}
M.~A. Fischler and R.~C. Bolles, ``Random sample consensus: A paradigm for
  model fitting with applications to image analysis and automated
  cartography,'' \emph{Commun. ACM}, vol.~24, no.~6, pp. 381--395, 1981.

\bibitem{liu2024egfs}
T.-R. Liu, H.-K. Yang, J.-M. Liu, C.-W. Huang, T.-C. Chiang, Q.~Kong,
  N.~Kobori, and C.-Y. Lee, ``Reprojection errors as prompts for efficient
  scene coordinate regression,'' in \emph{Proc. Eur. Conf. Comput. Vis.}, 2024.

\bibitem{chen2022epropnp}
H.~Chen, P.~Wang, F.~Wang, W.~Tian, L.~Xiong, and H.~Li, ``{EPro-PnP}:
  Generalized end-to-end probabilistic perspective-n-points for monocular
  object pose estimation,'' in \emph{Proc. IEEE/CVF Conf. Comput. Vis. Pattern
  Recognit. (CVPR)}, 2022, pp. 2781--2790.

\bibitem{fischer2020event}
T.~Fischer and M.~Milford, ``Event-based visual place recognition with
  ensembles of temporal windows,'' \emph{IEEE Robot. Autom. Lett.}, vol.~5,
  no.~4, pp. 6924--6931, 2020.

\bibitem{lee2021eventvlad}
A.~J. Lee and A.~Kim, ``{EventVLAD}: Visual place recognition with
  reconstructed edges from event cameras,'' in \emph{Proc. IEEE/RSJ Int. Conf.
  Intell. Robots Syst. (IROS)}, 2021, pp. 2247--2252.

\bibitem{fischer2022many}
T.~Fischer and M.~Milford, ``How many events do you need? {Event}-based visual
  place recognition using sparse but varying pixels,'' \emph{IEEE Robot. Autom.
  Lett.}, vol.~7, no.~4, pp. 12\,275--12\,282, 2022.

\bibitem{kong2022event}
D.~Kong, Z.~Fang, K.~Hou, H.~Li, J.~Jiang, S.~Coleman, and D.~Kerr,
  ``{Event-VPR}: End-to-end weakly supervised deep network architecture for
  visual place recognition using event-based vision sensor,'' \emph{IEEE Trans.
  Instrum. Meas.}, vol.~71, pp. 1--18, 2022.

\bibitem{nguyen2019real}
A.~Nguyen, T.-T. Do, D.~G. Caldwell, and N.~G. Tsagarakis, ``Real-time {6DOF}
  pose relocalization for event cameras with stacked spatial {LSTM} networks,''
  in \emph{Proc. IEEE/CVF Conf. Comput. Vis. Pattern Recognit. Workshops},
  2019, pp. 1638--1645.

\bibitem{jin20216}
Y.~Jin, L.~Yu, G.~Li, and S.~Fei, ``A {6-DOFs} event-based camera
  relocalization system by {CNN-LSTM} and image denoising,'' \emph{Expert Syst.
  Appl.}, vol. 170, p. 114535, 2021.

\bibitem{lin20226}
H.~Lin, M.~Li, Q.~Xia, Y.~Fei, B.~Yin, and X.~Yang, ``{6-DoF} pose
  relocalization for event cameras with entropy frame and attention networks,''
  in \emph{Proc. ACM SIGGRAPH Int. Conf. Virtual-Reality Continuum Appl. Ind.
  (VRCAI)}, 2022, pp. 1--8.

\bibitem{ren2024simple}
H.~Ren, J.~Zhu, Y.~Zhou, H.~Fu, Y.~Huang, and B.~Cheng, ``A simple and
  effective point-based network for event camera {6-DOFs} pose
  relocalization,'' in \emph{Proc. IEEE/CVF Conf. Comput. Vis. Pattern
  Recognit. (CVPR)}, 2024, pp. 18\,112--18\,121.

\bibitem{yuan2024evit}
R.~Yuan, T.~Liu, Z.~Dai, Y.-F. Zuo, and L.~Kneip, ``{EVIT}: Event-based
  visual-inertial tracking in semi-dense maps using windowed nonlinear
  optimization,'' in \emph{Proc. IEEE/RSJ Int. Conf. Intell. Robots Syst.
  (IROS)}, 2024, pp. 10\,656--10\,663.

\bibitem{zuo2024cross}
Y.-F. Zuo, W.~Xu, X.~Wang, Y.~Wang, and L.~Kneip, ``Cross-modal semidense
  {6-DOF} tracking of an event camera in challenging conditions,'' \emph{IEEE
  Trans. Robot.}, vol.~40, pp. 1600--1616, 2024.

\bibitem{gallego2017event}
G.~Gallego, J.~E.~A. Lund, E.~Mueggler, H.~Rebecq, T.~Delbruck, and
  D.~Scaramuzza, ``Event-based, {6-DOF} camera tracking from photometric depth
  maps,'' \emph{IEEE Trans. Pattern Anal. Mach. Intell.}, vol.~40, no.~10, pp.
  2402--2412, 2018.

\bibitem{bryner2019event}
S.~Bryner, G.~Gallego, H.~Rebecq, and D.~Scaramuzza, ``Event-based, direct
  camera tracking from a photometric {3D} map using nonlinear optimization,''
  in \emph{Proc. IEEE Int. Conf. Robot. Autom. (ICRA)}, 2019, pp. 325--331.

\bibitem{liu2024gs}
T.~Liu, R.~Yuan, Y.~Ju, X.~Xu, J.~Yang, X.~Meng, X.~Lagorce, and L.~Kneip,
  ``{GS-EVT}: Cross-modal event camera tracking based on {Gaussian}
  splatting,'' in \emph{Proc. IEEE Int. Conf. Robot. Autom. (ICRA)}, 2025, pp.
  4587--4593.

\bibitem{ta2023l2e}
K.~Ta, D.~Bruggemann, T.~Br{\"o}dermann, C.~Sakaridis, and L.~Van~Gool,
  ``{L2E}: Lasers to events for {6-DoF} extrinsic calibration of {LiDARs} and
  event cameras,'' in \emph{Proc. IEEE Int. Conf. Robot. Autom. (ICRA)}, 2023,
  pp. 11\,425--11\,431.

\bibitem{xing2023target}
W.~Xing, S.~Lin, L.~Yang, and J.~Pan, ``Target-free extrinsic calibration of
  event-{LiDAR} dyad using edge correspondences,'' \emph{IEEE Robot. Autom.
  Lett.}, vol.~8, no.~7, pp. 4020--4027, 2023.

\bibitem{lin2023e2pnet}
X.~Lin, C.~Qiu, Z.~Cai, S.~Shen, Y.~Zang, W.~Liu, X.~Bian, M.~M{\"u}ller, and
  C.~Wang, ``{E2PNet}: Event to point cloud registration with spatio-temporal
  representation learning,'' in \emph{Proc. Adv. Neural Inf. Process. Syst.
  (NeurIPS)}, vol.~36, 2023, pp. 18\,076--18\,089.

\bibitem{brachmann2019ngransac}
E.~Brachmann and C.~Rother, ``Neural-guided {RANSAC}: Learning where to sample
  model hypotheses,'' in \emph{Proc. IEEE/CVF Int. Conf. Comput. Vis. (ICCV)},
  2019, pp. 4322--4331.

\bibitem{teed2021droid}
Z.~Teed and J.~Deng, ``{DROID-SLAM}: Deep visual {SLAM} for monocular, stereo,
  and {RGB-D} cameras,'' in \emph{Proc. Adv. Neural Inf. Process. Syst.
  (NeurIPS)}, vol.~34, 2021, pp. 16\,558--16\,569.

\bibitem{chaney2023m3ed}
K.~Chaney, F.~Cladera, Z.~Wang, A.~Bisulco, M.~A. Hsieh, C.~Korpela, V.~Kumar,
  C.~J. Taylor, and K.~Daniilidis, ``{M3ED}: Multi-robot, multi-sensor,
  multi-environment event dataset,'' in \emph{Proc. IEEE/CVF Conf. Comput. Vis.
  Pattern Recognit. Workshops (CVPRW)}, 2023, pp. 4016--4023.

\bibitem{gehrig2021dsec}
M.~Gehrig, W.~Aarents, D.~Gehrig, and D.~Scaramuzza, ``{DSEC}: A stereo event
  camera dataset for driving scenarios,'' \emph{IEEE Robot. Autom. Lett.},
  vol.~6, no.~3, pp. 4947--4954, 2021.

\bibitem{PoseLib}
V.~Larsson and contributors, ``{PoseLib} -- minimal solvers for camera pose
  estimation,'' \url{https://github.com/PoseLib/PoseLib}, 2020.

\bibitem{sattler2019ransaclib}
T.~Sattler \emph{et~al.}, ``{RansacLib} -- a template-based {*SAC}
  implementation,'' \url{https://github.com/tsattler/RansacLib}, 2019.

\end{thebibliography}
\end{document}